\documentclass[conference]{IEEEtran}
\IEEEoverridecommandlockouts
\usepackage{cite}
\usepackage{amsmath,amssymb,amsfonts}
\usepackage{algorithmic}
\usepackage{graphicx}
\graphicspath{{figures/}}
\usepackage{textcomp}
\usepackage{xcolor}
\usepackage{float}
\usepackage{diagbox}
\usepackage{multirow}
\usepackage{subfig}

\def\BibTeX{{\rm B\kern-.05em{\sc i\kern-.025em b}\kern-.08em
    T\kern-.1667em\lower.7ex\hbox{E}\kern-.125emX}}

\begin{document}

\title{Visual Cue Integration for Small Target Motion Detection in Natural Cluttered Backgrounds
\thanks{This work was supported in part by EU FP7 Project HAZCEPT under Grant 318907, in part by HORIZON 2020 project STEP2DYNA under Grant 691154, in part by HORIZON 2020 project ULTRACEPT under Grant 778062, and in part by the National Natural Science Foundation of China under Grant 11771347.}
}

\author{
	\IEEEauthorblockN{Hongxin Wang\IEEEauthorrefmark{1},  Jigen Peng\IEEEauthorrefmark{2}, Qinbing Fu\IEEEauthorrefmark{1}\IEEEauthorrefmark{3}, Huatian Wang\IEEEauthorrefmark{1}, Shigang Yue\IEEEauthorrefmark{1}\IEEEauthorrefmark{3}} 
	
	\IEEEauthorblockA{\IEEEauthorrefmark{1} School of Computer Science, University of Lincoln, Lincoln, UK} 
	\IEEEauthorblockA{\IEEEauthorrefmark{2} School of Mathematics and Information Science, Guangzhou University, Guangzhou, China} 
	\IEEEauthorblockA{\IEEEauthorrefmark{3} Machine Life and Intelligence Research Center, Guangzhou University, Guangzhou, China} 
	howang@lincoln.ac.uk, jgpeng@gzhu.edu.cn, \{qifu, hwang, syue\}@lincoln.ac.uk
}


\maketitle

\begin{abstract}
The robust detection of small targets against cluttered background is important for future artificial visual systems in searching and tracking applications. The insects' visual systems have demonstrated excellent ability to avoid predators, find prey or identify conspecifics -- which always appear as small dim speckles in the visual field. Build a computational model of the insects' visual pathways could provide effective solutions to detect small moving targets. Although a few visual system models have been proposed, they only make use of small-field visual features for motion detection and their detection results often contain a number of false positives. To address this issue, we develop a new visual system model for small target motion detection against cluttered moving backgrounds. Compared to the existing models, the small-field and wide-field visual features are separately extracted by two motion-sensitive neurons to detect small target motion and background motion. These two types of motion information are further integrated to filter out false positives. Extensive experiments showed that the proposed model can outperform the existing models in terms of detection rates.
\end{abstract}


\section{Introduction}
As processing power increases exponentially, and as sensors becomes less costly and more reliable, robots have shown great potential in reshaping human life in the future \cite{yue2006collision,fu2018shaping,hu2016bio,wang2018model,wang2018feedback}. Intelligent robots embedded with artificial visual systems will be able to cope with dynamic visual worlds in real-time and perform required tasks without human intervention \cite{yue2006bio,Hu2016A,yue2013redundant,yue2017modeling,song2018fast}. Among a number of visual functionalities, detecting objects of interest in the distance and early could help a robot achieve dominant position in competition, defence, and survive. However, artificial visual systems are still far from acceptable to robustly and cheaply detect moving objects in the distance against cluttered natural backgrounds.

In the visual world, detecting object motion which is far away from the observer, often means dealing with small dim speckles in the field of view. The difficulty of small target motion detection is reflected in: first, the sizes of small targets may vary from one pixel to a few pixels, whereas other physical characteristics, such as color, shape and texture, are difficult to recognize and cannot be used for motion detection. Second, the natural background is extremely cluttered and contains a number of small-target-like features. In addition, free motion of camera would bring further difficulties to small target motion detection.

For animals, such as insects, the ability to detect small targets against cluttered backgrounds is important, serving to search for mates and track prey. Evolved over millions of years, the small target detection visual systems in insects are both efficient and reliable. As an example, dragonflies can pursue small flying insects with successful capture rates as high as $97 \%$ relying on their well evolved visual systems \cite{olberg2000prey}. The exquisite sensitivity of insects for small moving targets is coming from a class of specific neurons, called small target motion detectors (STMDs) \cite{o1993feature}. The STMD neurons give peak responses to small targets subtending $1-3^{\circ}$ of the visual field, with no response to large bars (typically $>10^{\circ}$) or to background movements represented by wide-field  grating stimuli \cite{nordstrom2012neural}. This makes the STMD an ideal template to develop specialized artificial visual systems for small target motion detection.

The electrophysiological knowledge about the STMD neuron revealed in the past few decades, makes it possible to propose quantitative models. Wiederman \emph{et al.} \cite{wiederman2008model} presented an elementary STMD (ESTMD) to account for size selectivity of the STMD neurons. However, the ESTMD did not consider direction selectivity and is unable to estimate motion direction of small targets. To address this issue, some directionally selective STMD models have been developed, including two hybrid models \cite{wiederman2013biologically,bagheri2017performance}, and directionally selective STMD (DSTMD) \cite{wang2018directionally}. These STMD-based models take advantage of small-field visual features for small target motion detection, while ignore other visual cues, such as wide-field features. Due to this, the models cannot discriminate small target motion from  false positive background motion, and their detection results often contain a large number of noises. In order to eliminate the background false positives, the wide-field features should not be disregarded, which can be combined with the small-field cues in small target motion discrimination.

In the insects' visual systems, the motion of wide-field features can elicit strong responses of an identified interneuron called lobula plate tangential cell (LPTC) \cite{borst1995mechanisms}. In the further research \cite{nicholas2018integration}, Nicholas \emph{et al.} assert that the wide-field motion and small target motion which are separately detected by the LPTCs and STMDs, are integrated in target-selective descending neurons (TSDNs). More precisely, the responses of the STMDs are largely suppressed by the LPTCs when the small object and background move at the similar velocities. These biological findings provide a solution for designing artificial visual systems to filter out false positives in small target motion detection.

Inspired by the above biological findings, this paper proposes a visual system model (TSDN) for small target motion detection in cluttered moving backgrounds. The visual system is composed of an STMD subsystem for small target motion detection and an LPTC subsystem for wide-field motion perception. The outputs of the STMD and LPTC are integrated to discriminate small targets from background false positives. The rest of this paper is organized as follows. In Section \ref{Methods}, we introduce our proposed visual system model. Section \ref{Results} provides extensive performance evaluation as well as comparisons against the existing models. Finally, we conclude this paper in Section \ref{Conclusion}.

\section{Methods}
\label{Methods}

\begin{figure}[t!]
	\centering
	\includegraphics[width=0.45\textwidth]{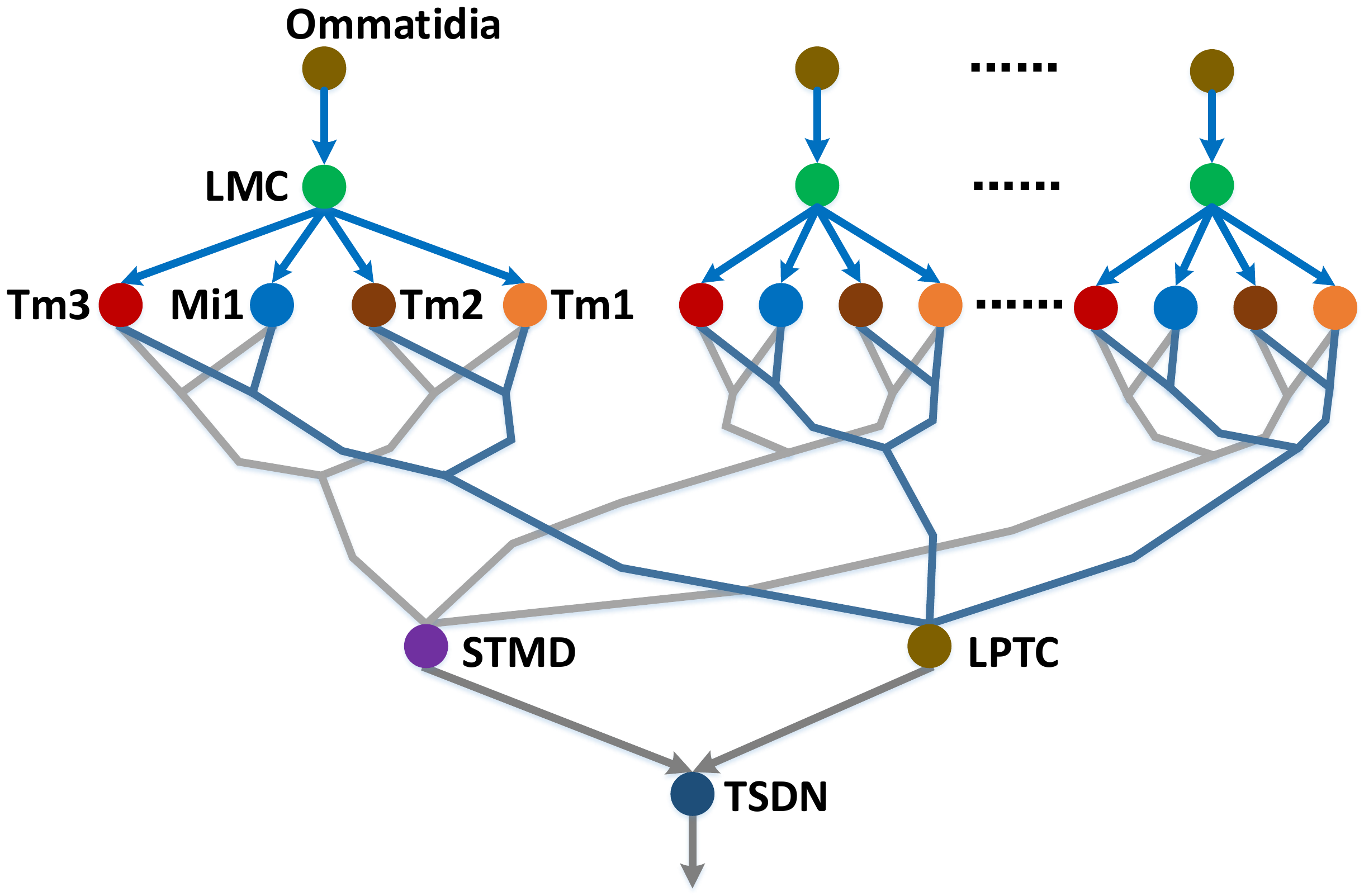}
	\caption{Wiring sketch of the proposed visual system where each colored node denotes a neuron. For clear illustration, only one STMD, LPTC and TSDN are presented here.}
	\label{Schematic-of-Visual-System}
\end{figure}

As illustrated in Fig. \ref{Schematic-of-Visual-System}, our developed visual system is composed of ommatidia, large monopolar cells (LMCs) \cite{harrison2003low}, medulla neurons (including Tm1, Tm2, Tm3, and Mi1) \cite{behnia2015visual}, STMDs, LPTCs and TSNDs. Luminance signals are received and smoothed by the ommatidia, then applied to the LMCs where lumninace changes of each pixel over time are extracted. These luminance change signals are further separated into luminance increase and decrease components by the medulla neurons, which are fed into the STMDs and LPTCs for detecting small target motion and wide-field motion, respectively. The TSDNs integrate the extracted motion information to filter out the background false positives. The schematic illustration of the proposed visual system model is presented in Fig. \ref{Schematic-of-the-Proposed-TSDN-Model} which will be elaborated in the following.

\subsection{Ommatidia}
The first layer of the developed visual system is the ommatidia arranged in a matrix; the luminance of each pixel in the input image is captured and smoothed by each ommatidium which is modelled as a spatial Gaussian filter. Let $I(x,y,t) \in \mathbb{R}$ denote the input image sequence, where $x,y$ and $t$ are spatial and temporal field positions. The output of an ommatidium $P(x,y,t)$ is given by,
\begin{equation}
P(x,y,t) =  \iint I(u,v,t)G_{\sigma_1}(x-u,y-v)dudv
\label{Photoreceptors-Gaussian-Blur}
\end{equation}
where $G_{\sigma_1}(x,y)$ is a Gaussian function, defined as
\begin{equation}
G_{\sigma_1}(x,y)= \frac{1}{2\pi\sigma_1^2}\exp(-\frac{x^2+y^2}{2\sigma_1^2}).
\label{Photoreceptors-Gauss-blur-Kernel}
\end{equation}

\subsection{Large Monopolar Cells}

\begin{figure*}[t!]
	\centering
	\includegraphics[width=1.0\textwidth]{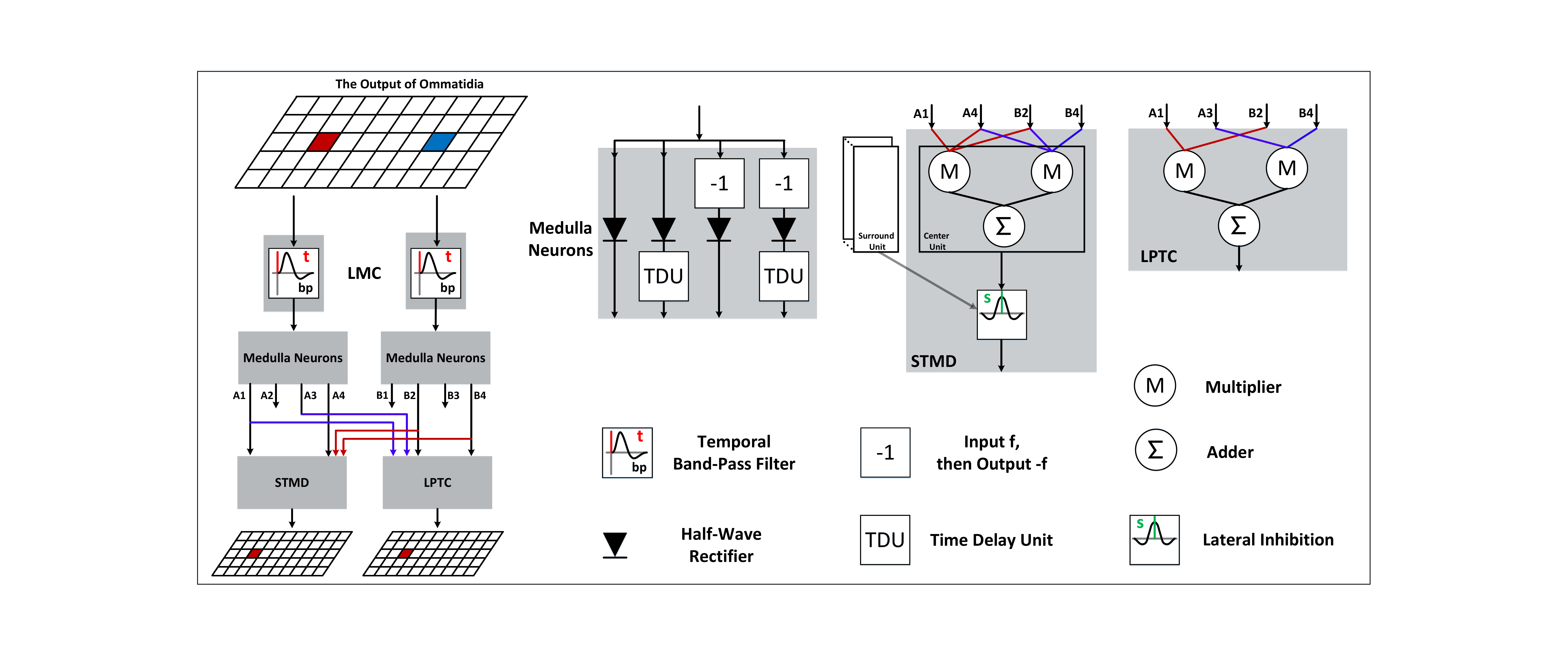}
	\caption{Schematic illustration of the proposed visual system model. For clear illustration, only one STMD and LPTC are presented here. However, these types of neurons are all arranged in matrix form in the proposed visual system model.}
	\label{Schematic-of-the-Proposed-TSDN-Model}
\end{figure*}

As shown in Fig. \ref{Schematic-of-the-Proposed-TSDN-Model}, the output of the ommatidia forms the input to LMCs in the next layer. Each LMC is modelled as a temporal band-pass filter to extract luminance changes over time caused by motion. The impulse response of the band-pass filter $H(t)$ is defined as the difference of two Gamma kernels, then the output of each LMC $L(x,y,t)$ can be given by,
\begin{align}
L(x,y,t) &= \int P(x,y,s)H(t-s) ds \\
H(t) &= \Gamma_{n_1,\tau_1}(t) - \Gamma_{n_2,\tau_2}(t)
\label{BPF-Para}
\end{align}
where $\Gamma_{n,\tau}(t)$ stands for the Gamma kernel \cite{de1991theory}, defined as 
\begin{equation}
\Gamma_{n,\tau}(t) = (nt)^n \frac{\exp(-nt/\tau)}{(n-1)!\cdot \tau^{n+1}}.
\end{equation}
where $n$ and $\tau$ represents the order and time constant of Gamma kernel.

\subsection{Medulla Neurons}

As can be seen from Fig. \ref{Schematic-of-the-Proposed-TSDN-Model}, the output of LMCs $L(x,y,t)$ is applied to the medulla neurons including Tm1, Tm2, Tm3, and Mi1, which constitute four parallel signal-processing channels. The Tm3 and Tm2 are modelled as half-wave rectifiers to separate $L(x,y,t)$ into luminance increase and decrease components. Let $S^{\text{Tm3}}(x,y,t)$ and $S^{\text{Tm2}}(x,y,t)$ denote the output of the Tm3 and Tm2, respectively, then they are given by
\begin{align}
S^{\text{Tm3}}(x,y,t) &= [L(x,y,t)]^{+}  \label{Tm3-Output} \\
S^{\text{Tm2}}(x,y,t) &= [-L(x,y,t)]^{+} \label{Tm2-Output}
\end{align}
where $[x]^+$ denotes $\max (x,0)$. The Mi1 and Tm1 further temporally delay $S^{\text{Tm3}}(x,y,t)$ and $S^{\text{Tm2}}(x,y,t)$ by convolving them with a Gamma kernel. That is, 
\begin{align}
S_{{(n,\tau)}}^{\text{Mi1}}(x,y,t) &= \int [L(x,y,s)]^{+} \cdot \Gamma_{n,\tau}(t-s) ds \label{Mi1-Output}\\
S_{{(n,\tau)}}^{\text{Tm1}}(x,y,t) &= \int [-L(x,y,s)]^{+} \cdot  \Gamma_{n,\tau}(t-s) ds \label{Tm1-Output}
\end{align}
where $S_{{(n,\tau)}}^{\text{Mi1}}(x,y,t)$ and $S_{{(n,\tau)}}^{\text{Tm1}}(x,y,t)$ represent the outputs of the Mi1 and Tm1, respectively; $n$ and $\tau$ are the order and time constant of the Gamma kernel, which separately determine the time-delay order and length.

\subsection{Small Target Motion Detectors}

As illustrated in Fig. \ref{Schematic-of-the-Proposed-TSDN-Model}, the outputs of the medulla neurons are collected by the STMDs for small target motion detection against cluttered moving backgrounds. The existing DSTMD model \cite{wang2018directionally} is adopted to describe the STMDs, where the output of each STMD $D(x,y,t,\theta)$ is defined by the multiplication of medulla neural outputs from two different pixels to produce strong responses to moving objects. That is,
\begin{equation}
\begin{split}
D(x, & y,t,\theta) = S^{\text{Tm3}}(x,y,t) \cdot \Big \{S^{\text{Tm1}}_{{(n_{_4},\tau_{_4})}}(x,y,t)\\  &+S^{\text{Mi1}}_{{(n_{_3},\tau_{_3})}}(x'(\theta),y'(\theta),t)\Big\} \cdot S^{\text{Tm1}}_{{(n_{_5},\tau_{_5})}}(x'(\theta),y'(\theta),t)
\label{DSTMD-Signal-Correlation}
\end{split}
\end{equation}
where $\theta$ denotes the preferred direction of the STMD neuron; $(x'(\theta),y'(\theta))$ is defined as 
\begin{equation}
\begin{split}
x'(\theta) &= x+\alpha_1\cos\theta \\ 
y'(\theta) &= y + \alpha_1\sin\theta
\end{split}
\label{DSTMD-Signal-Correlation-Distance}
\end{equation}
where $\alpha_1$ is a constant.

In order to suppress the responses to the large moving objects, the $D(x,y,t,\theta)$ is laterally inhibited by convolving with an inhibition kernel $W_s(x,y)$. That is,
\begin{equation}
E(x,y,t,\theta) = \iint D(u,v,t,\theta) W_s(x-u,y-v) du dv
\label{DS-STMD-Lateral-Inhibition}
\end{equation}
where $E(x,y,t,\theta)$ represents the inhibited signal; the inhibition kernel $W_s(x,y)$ is defined as
\begin{align}
W_s(x,y) &= A \cdot [g(x,y)]^{+} + B \cdot [g(x,y)]^{-}  \label{Inhibition-Kernel-W2-1}\\
g(x,y)  &= G_{\sigma_2}(x,y) - e \cdot G_{\sigma_3}(x,y) - \rho
\label{DSTMD-Lateral-Inhibition-Kernel-W2-2}
\end{align}
where $[x]^+$ and $[x]^-$ respectively denote $\max (x,0)$ and $\min (x,0)$; $A$, $B$, $e$ and $\rho$ are constant. 

The output of STMDs $E(x,y,t,\theta)$ can be used to determine the positions of small moving targets by comparing it with a threshold $\beta$. Specifically, if $E(x,y,t,\theta)$ is higher than $\beta$, then we believe that a small object moving along direction $\theta$ is located at pixel $(x,y)$ and time $t$.

\subsection{Lobula Plate Tangential Cells}

Although the LPTC also gather the outputs of medulla neurons, it serves to detect wide-field motion rather than small target motion. The LPTC is  modelled by the existing two-quadrant-detector (TQD) \cite{eichner2011internal,wang2018improved}, and its output $F(x,y,t,\psi)$ is defined as
\begin{equation}
\begin{split}
F(x,&y,t,\psi) =  S^{\text{Tm3}}(x,y,t) \cdot S^{\text{Mi1}}_{{(n_{_6},\tau_{_6})}}(x'(\psi),y'(\psi),t) \\  
& + S^{\text{Tm2}}(x,y,t) \cdot S^{\text{Tm1}}_{{(n_{_6},\tau_{_6})}}(x'(\psi),y'(\psi),t)
\end{split}
\label{Classic-TQD-Signal-Correlation}
\end{equation}
where $\psi$ stands for the preferred direction of the LPTC neuron.  

The output of LPTCs $F(x,y,t,\psi)$ reveals the motion of normal-size objects embedded in the cluttered background such as trees, rocks and bushes. More precisely, for a threshold $\gamma$, if $F(x,y,t,\psi) > \gamma$, then we believe that a background object moving along direction $\psi$ is detected at pixel $(x,y)$ and time $t$.

\subsection{Target-Selective Descending Neurons}

The TSDN receives two types of neural outputs, including the output of STMDs $E(x,y,t,\theta)$ and the output of LPTCs $F(x,y,t,\psi)$. These neural outputs are integrated to filter out background false positives via the following two steps. 

{\textit{1) Background Motion Direction Estimation}:} The output of LPTCs $F(x,y,t,\psi)$ is used to estimate the motion direction of the background. The basic idea is using the motion direction of most background objects to represent that of the background. That is,
\begin{equation}
\Psi(t) = \arg \max_{\psi} \iint F(x,y,t,\psi) dx dy
\label{Background-Motion-Direction-Estimation}
\end{equation}
where $\Psi(t)$ denotes the motion direction of the background at time $t$.

{\textit{2) False Positive Elimination}:} As revealed in the biological research \cite{nicholas2018integration}, the output of STMDs is largely inhibited by the output of LPTCs when the target and background move in the same direction.  That is, the excitatory flow from the STMDs $E(x,y,t,\theta)$ and  inhibition from the LPTCs $F(x,y,t,\psi)$ are summed by the TSDNs using the following:
\begin{equation}
T(x,y,t,\theta) = E(x,y,t,\theta) - \alpha_2 F(x,y,t,\Psi(t))
\label{TSDNs-Outputs}
\end{equation}
where $T(x,y,t,\theta)$ refers to the output of TSDNs and $\alpha_2 > 0$, if $\theta = \Psi(t)$; otherwise, $\alpha_2 = 0$.

It is worthy to note that the false positives are often caused by the motion of objects embedded in the background, which means their motion directions are consistent with that of the background. In the (\ref{TSDNs-Outputs}), the responses of STMDs will deduct the outputs of LPTCs, if the motion direction of the detected object $\theta$ equals to that of the background $\Psi(t)$. That is, the responses to the false positives will decrease, resulting in the improvement of detection performances.

\subsection{Parameter Setting}
Parameters of the proposed visual system model are listed in Table \ref{Table-Parameter-FVS}. Based on the previous parameter analysis and test \cite{wang2018directionally,wang2018improved,wang2018feedback,wang2016bio}, the parameters are tuned manually to make the developed model satisfy the basic neural properties, which are mainly determined by target velocity and size. They will not be changed in the following experiments unless stated.

The proposed visual system model is written in Matlab (The MathWorks, Inc., Natick, MA). The computer used in the experiments is a standard laptop with a $2.50$GHz Intel Core i7 CPU and $16$GB DDR3 memory.

\begin{table}[t!]
	\renewcommand{\arraystretch}{1.3}
	\caption{Parameters of the proposed visual system model.}
	\label{Table-Parameter-FVS}
	\centering
	\begin{tabular}{cc}
		\hline
		Eq. & Parameters \\	
		\hline
		(\ref{Photoreceptors-Gaussian-Blur}) & $\sigma_1 = 1$ \\
		
		(\ref{BPF-Para}) & $n_1 = 2, \tau_1= 3, n_2 = 6,\tau_2 = 9$\\
		
		(\ref{DSTMD-Signal-Correlation}) & $n_3 = 3, \tau_3 = 15, n_4 = 5, \tau_4 = 25, n_5 = 8, \tau_5 = 40$ \\
		
		(\ref{DSTMD-Signal-Correlation-Distance}) & $\alpha_1 = 3$ \\
		
		(\ref{Inhibition-Kernel-W2-1}) & $A = 1, B = 3$ \\
		
		(\ref{DSTMD-Lateral-Inhibition-Kernel-W2-2}) & $\sigma_2 = 1.5, \sigma_3 = 3.0, e = 1, \rho = 0$ \\
		
		(\ref{Classic-TQD-Signal-Correlation}) & $n_3 = 5, \tau_3 = 15$ \\
		
		(\ref{TSDNs-Outputs}) & $\alpha_2 = 3.5$ \\

		\hline
	\end{tabular}
\end{table}

\section{Results}
\label{Results}

The developed visual system model is evaluated on a dataset which is produced by Vision Egg \cite{straw2008vision}. The image sequences in the dataset are all synthesized by using real natural background images and a computer-generated small target with different sizes, velocities and luminance. The video images are $500$ (in horizontal) by $250$ (in vertical) pixels and temporal sampling frequency is set as $1000$ Hz.

In order to quantitatively evaluate the detection performance, two metrics are defined as following \cite{gao2013infrared},
\begin{align}
D_R & = \frac{\text{number of true detections}}{\text{number of actual targets}} \\
F_A & = \frac{\text{number of false detections}}{\text{number of images}}
\end{align}
where $D_R$ and $F_A$ denote detection rate and false alarm rate, respectively. The detected result is considered correct if the pixel distance between the ground truth and the result is within a threshold ($5$ pixels).

\subsection{Differences of the STMD and LPTC}

\begin{figure}[t]
	\centering
	\includegraphics[width=0.20\textwidth]{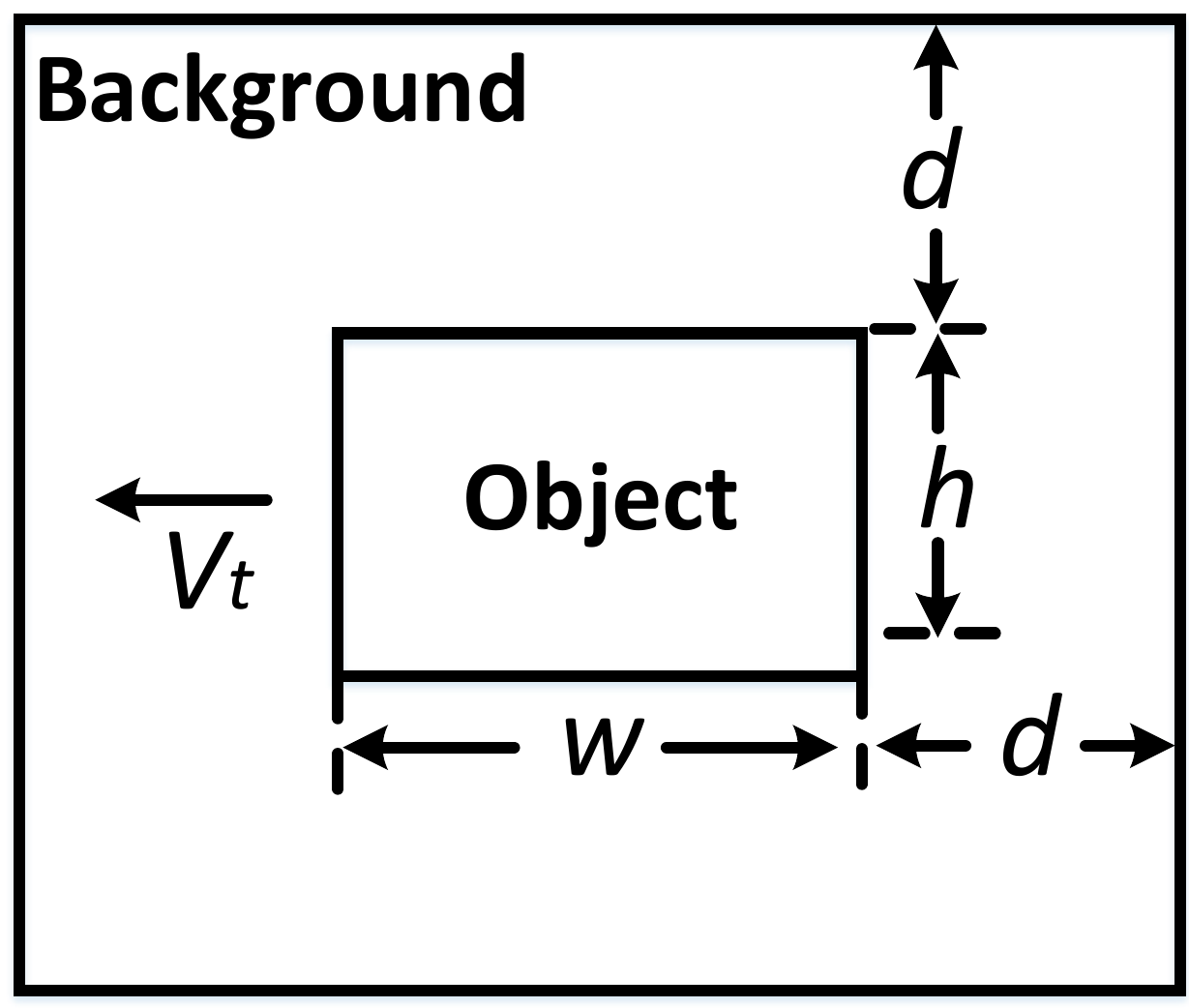}
	\caption{The external rectangle and neighboring background rectangle of an object. Arrow $V_T$ denotes the motion direction of the object. $w$ represents object width while $h$ stands for object height.}
	\label{The-External-Rectangle-and-Neighboring-Background-Rectangle-of-a-Small-Target}
\end{figure}

\begin{figure}[t]
	\vspace{-10pt}
	\centering
	\subfloat[]{\includegraphics[width=0.225\textwidth]{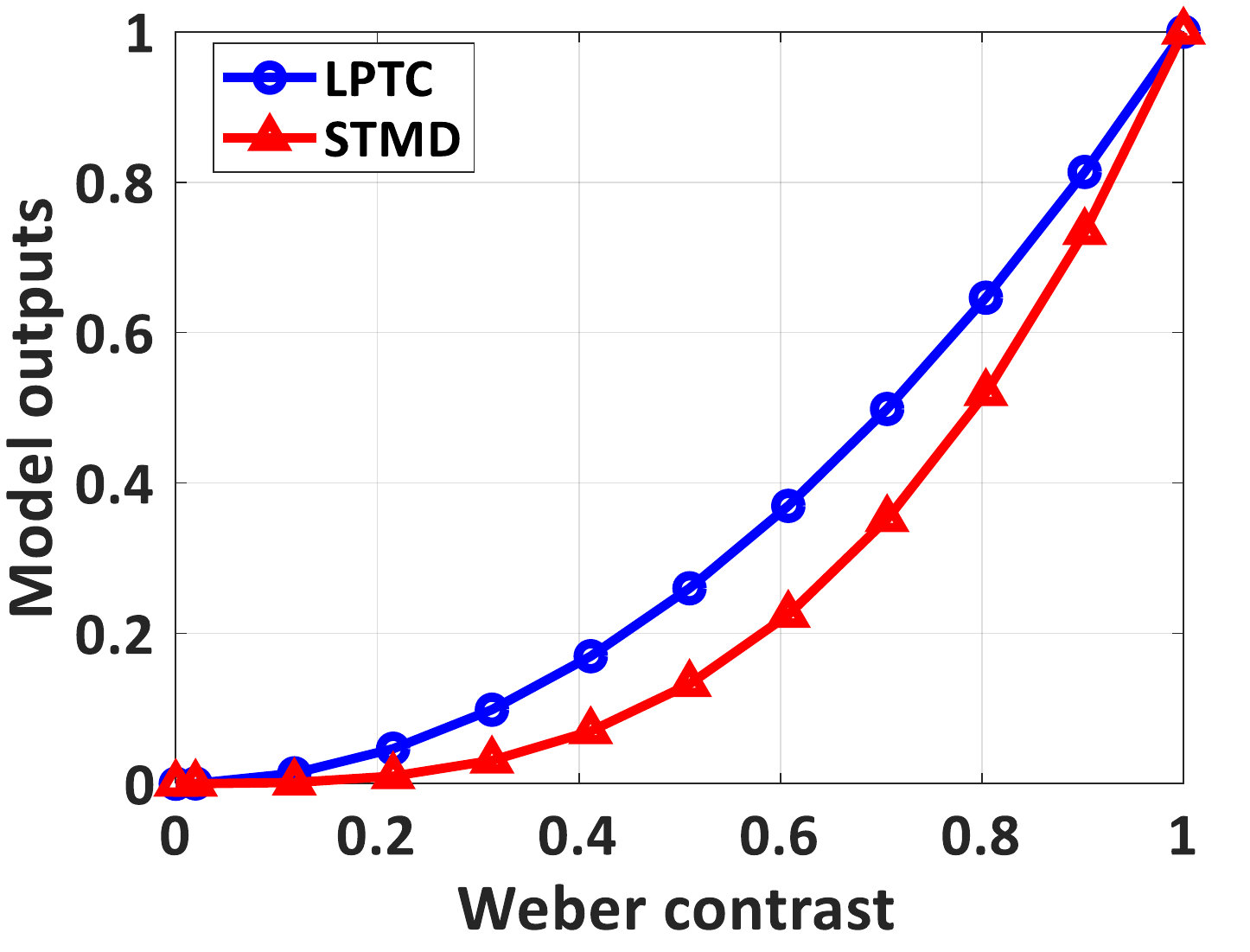}
		\label{Tuning-Properties-Contrast-LPTC-STMD}}
	\hfil
	\subfloat[]{\includegraphics[width=0.225\textwidth]{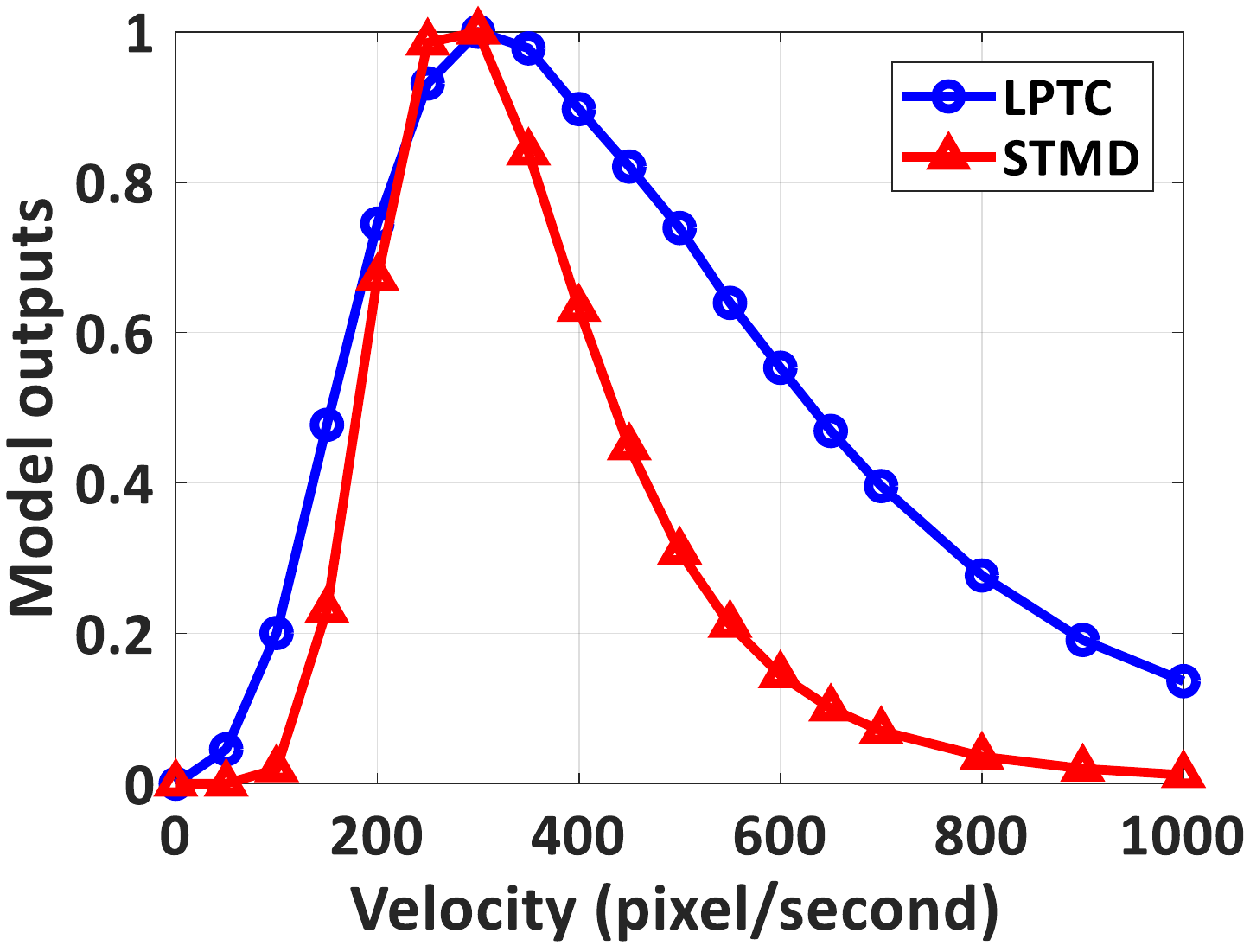}
		\label{Tuning-Properties-Velocity-LPTC-STMD}}
	\hfil
	\subfloat[]{\includegraphics[width=0.225\textwidth]{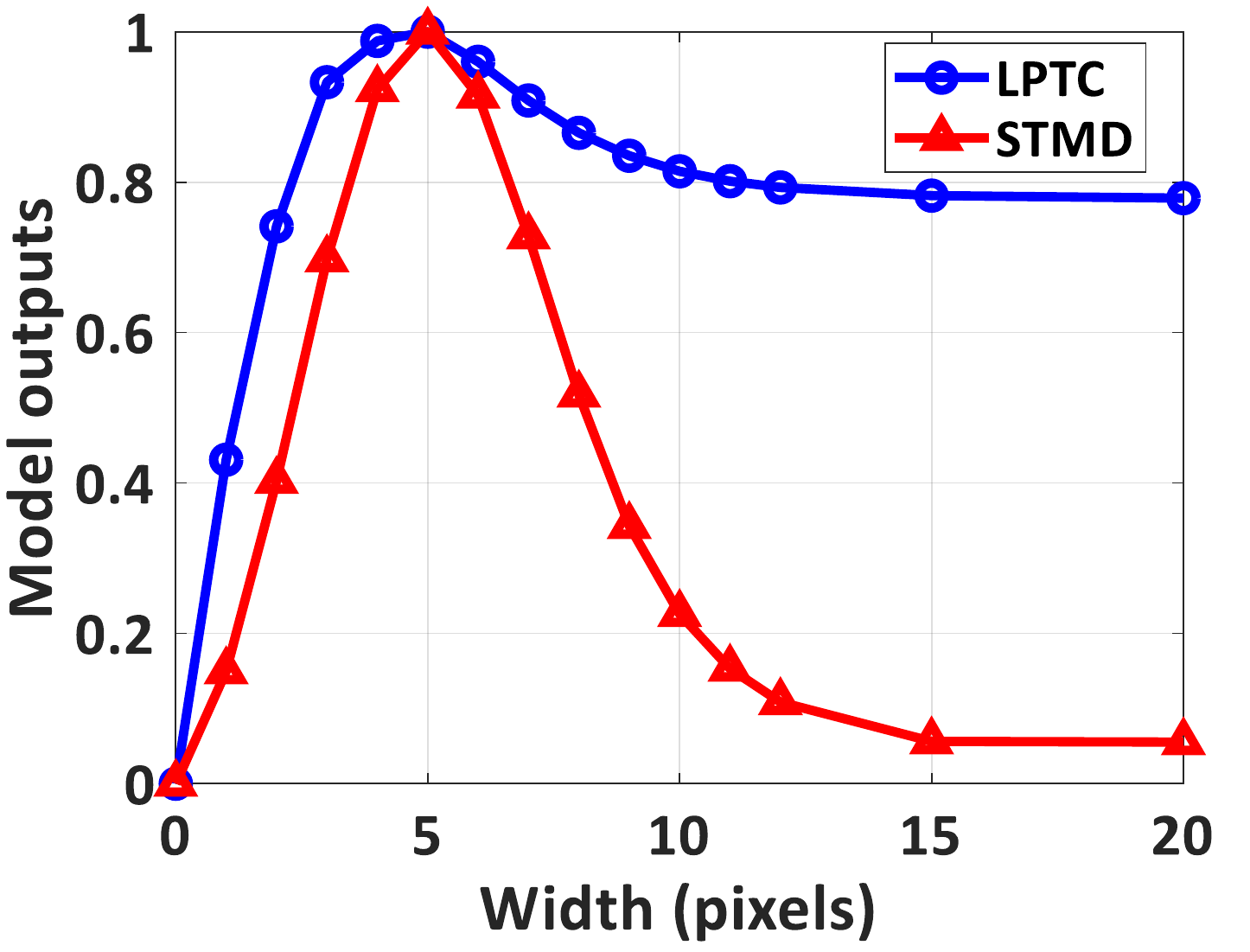}
		\label{Tuning-Properties-Width-LPTC-STMD}}
	\hfil
	\subfloat[]{\includegraphics[width=0.225\textwidth]{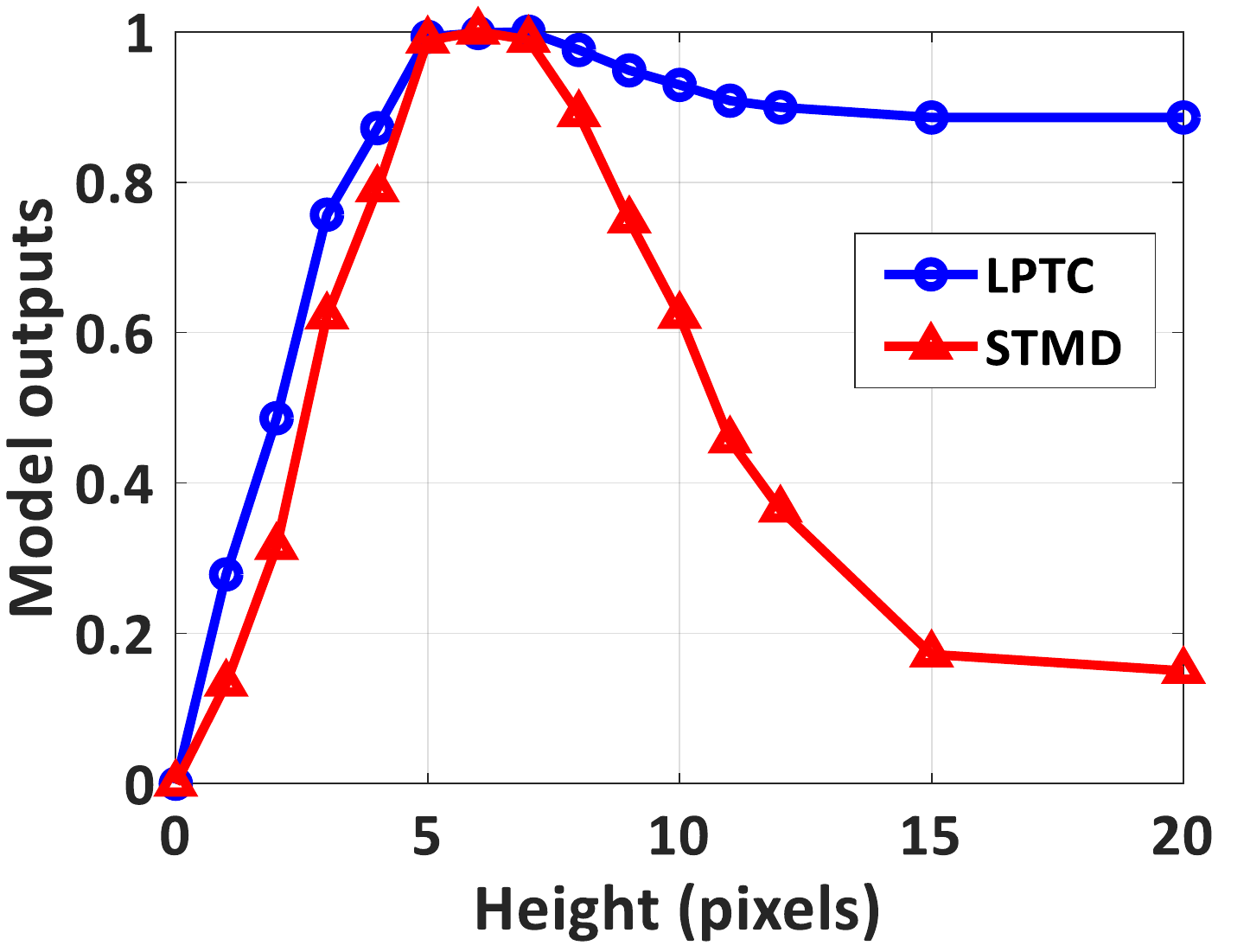}
		\label{Tuning-Properties-Height-LPTC-STMD}}
	
	\caption{STMD and LPTC outputs to moving objects with different Weber contrast, velocities, widths and heights. (a) Different Weber contrast. (b) Different velocities. (c) Different widths. (d) Different heights.}
	\label{Tuning-Properties-LPTC-STMD}
\end{figure}

In the proposed visual system model, the STMD and LPTC are applied to detect small target motion and background motion, respectively. To further demonstrate the differences between the STMD and LPTC, we compare their outputs to objects with different velocities, widths, heights and Weber contrast. As shown in Fig. \ref{The-External-Rectangle-and-Neighboring-Background-Rectangle-of-a-Small-Target}, width (or height) represents object length extended parallel (or orthogonal) to the motion direction. Weber contrast is defined by the following equation,
\begin{equation}
\text{Weber contrast} = \frac{|\mu_t - \mu_b|}{255}
\label{LDTB}
\end{equation}
where $\mu_t$ is the average pixel value of the object, while $\mu_b$ is the average pixel value in neighboring area around the object. If the size of a object is $w \times h$, the size of its background rectangle is $(w+2d)\times(h+2d)$, where $d$ is a constant which equals to $10$ pixels. The initial Weber contrast, velocity, width and height of the object are set as $1$, $250$ pixel/second, $5$ pixels and $5$ pixels, respectively. 

Fig. \ref{Tuning-Properties-LPTC-STMD}(a) shows the STMD and LPTC outputs with respect to the  Weber contrast. As can be seen, both the STMD and LPTC outputs increase as the increase of Weber contrast, until reach maximum at Weber contrast $=1$. This indicates that the higher Weber contrast of an object is, the easier it can be detected. Fig. \ref{Tuning-Properties-LPTC-STMD}(b) presents the two neural outputs with regard to the velocity of the moving object. Obviously, the STMD and LPTC outputs peak at an optimal velocity ($300$ pixel/s). These two neurons also exhibit high responses to the objects whose velocities range from $100$ to $600$ pixel/s. Fig. \ref{Tuning-Properties-LPTC-STMD}(c) and (d) display the outputs of the STMD and LPTC when changing the width and height of the object. As it is shown, both the STMD and LPTC outputs have a local maximum at width $= 5$ (or height $= 5$). After reaching the local maximum, the STMD output decreases significantly as the increase of target width (or height), and tends to be stable around $0.1$ (or $0.2$ for height). In contrast, the LPTC output has a slight drop and stabilizes at $0.8$ (or $0.9$ for height). The above results indicate that the STMD prefers small moving objects whose widths and heights are smaller than $10$ pixels, while the LPTC shows little preference for the target's width and height and can detect moving objects with normal sizes in the backgrounds.

\subsection{Effectiveness of the TSDN}

\begin{figure}[t]
	\centering
	\subfloat[]{\includegraphics[width=0.35\textwidth]{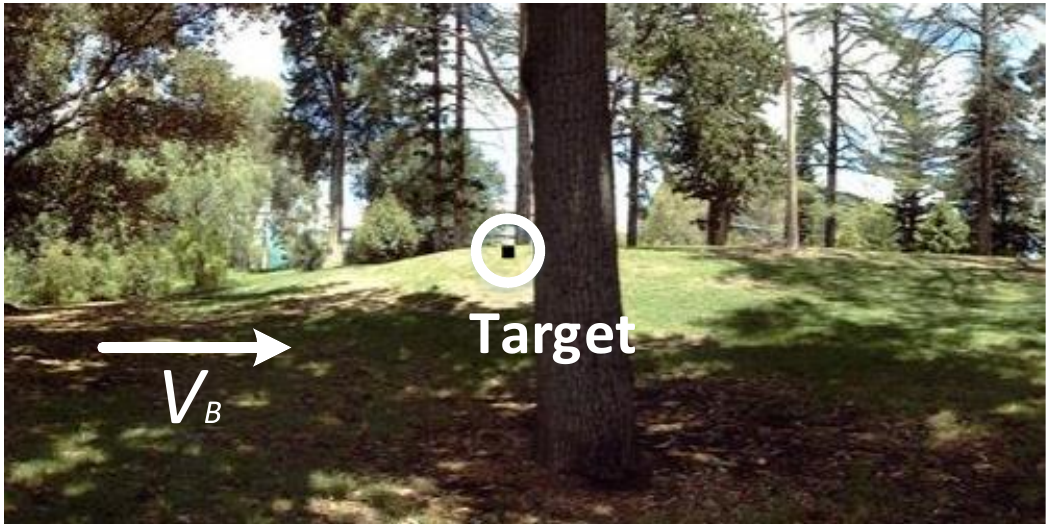}
		\label{Example-Input-Image}}
	\hfil
	\subfloat[]{\includegraphics[width=0.5\textwidth]{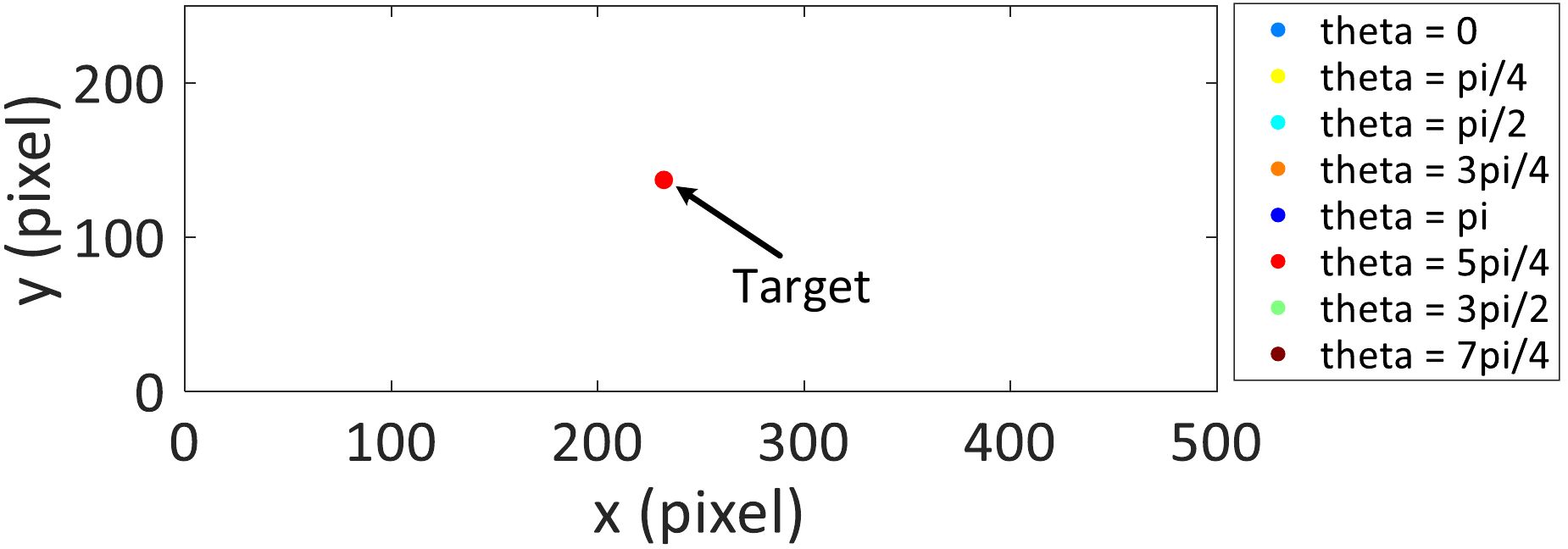}
		\label{Detection-Result-Ground-Truth}}
	
	\caption{(a) Representative frame of the input image sequence. A small target (the small black block) highlighted by the circle, is moving against the cluttered background. The cluttered background is also moving from left to right where arrow $V_B$ denotes the background motion direction. (b) The position of the small target at time $t=1000$ ms, i.e., ground truth. In this subplot, color represents motion direction $\theta$ of the small target.}
\label{Curvilinear-Motion-Original-Image-and-Target-Trace}
\end{figure}

\begin{figure*}[!t]
	\centering
	\subfloat[]{\includegraphics[width=0.425\textwidth]{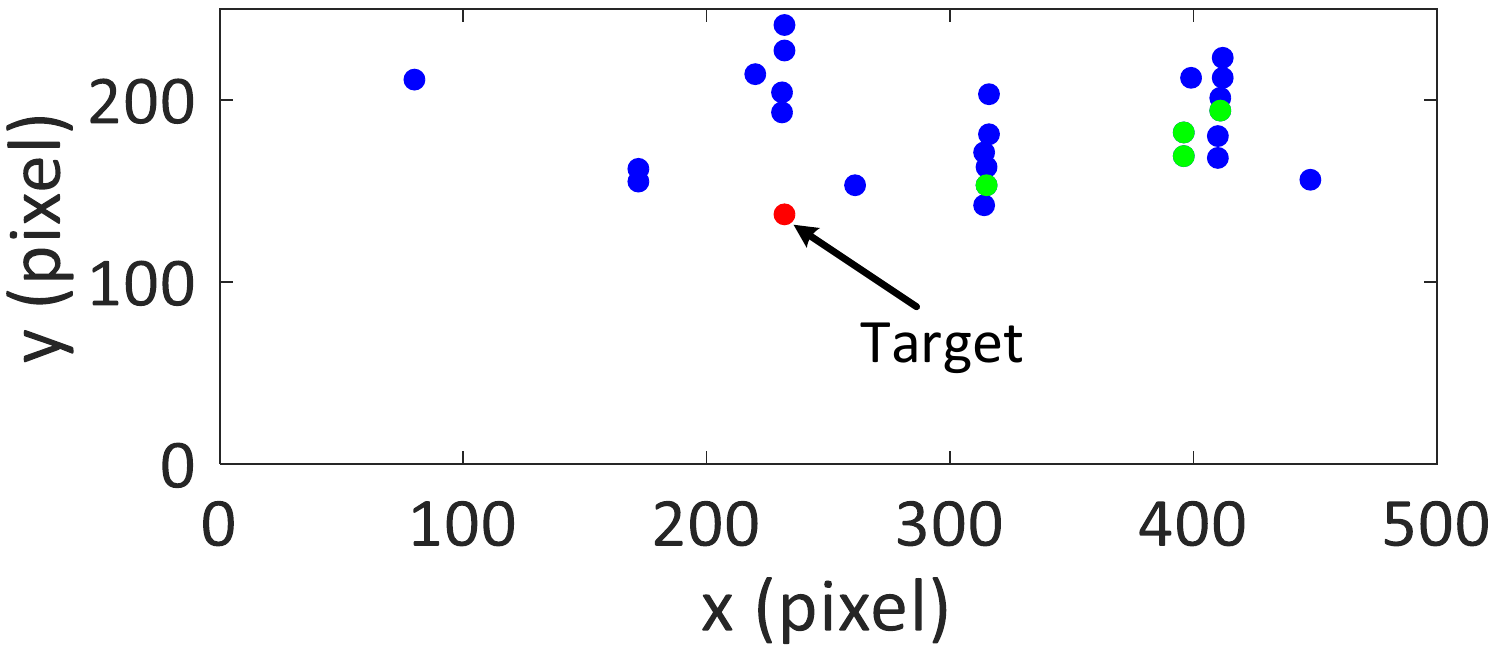}
		\label{Detection-Result-DSTMD}}
	\hfil
	\subfloat[]{\includegraphics[width=0.425\textwidth]{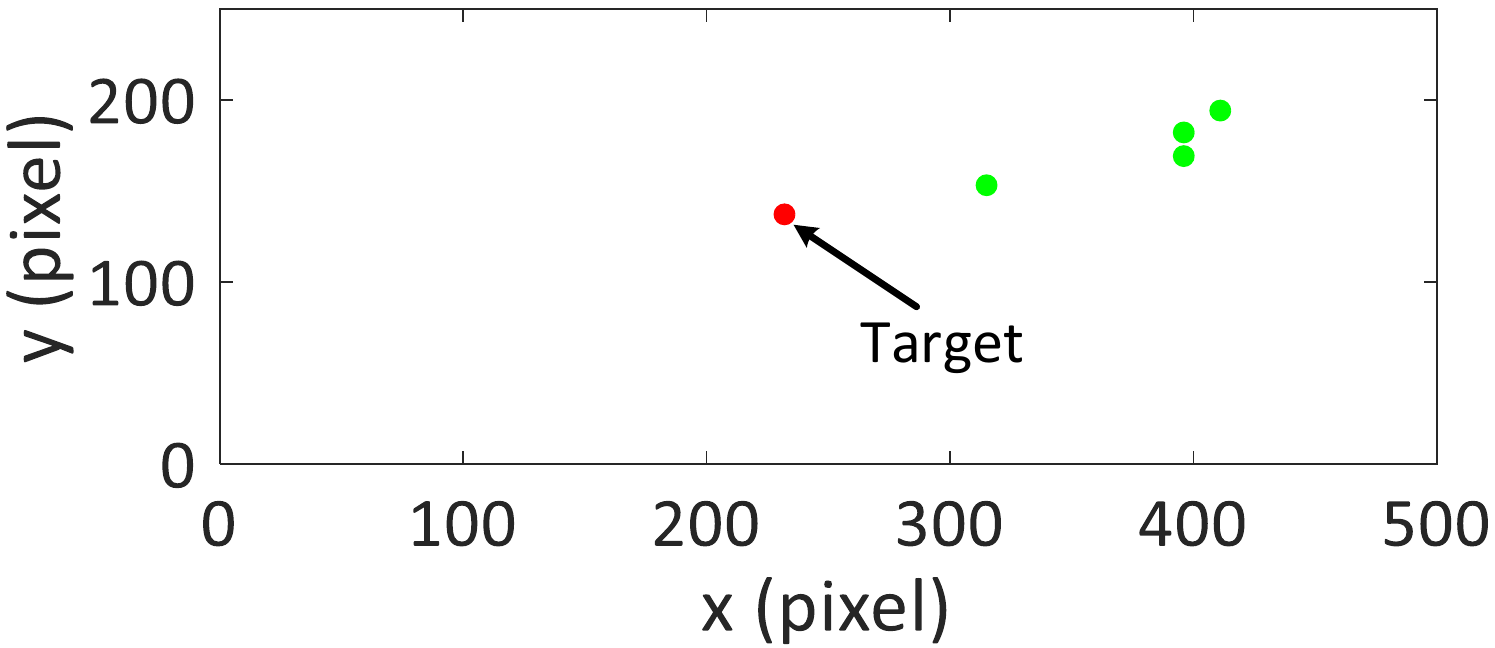}
		\label{Detection-Result-TSDN}}
	\caption{(a)-(b) Detection results of the STMDs $E(x,y,t,\theta)$ and TSDNs $T(x,y,t,\theta)$, respectively, where the detection threshold $\beta$ is set as $150$.}
	\label{Detection-Result-TSDN-DSTMD}
\end{figure*}

\begin{figure*}[!t]
	\centering
	\subfloat[]{\includegraphics[width=0.3\textwidth]{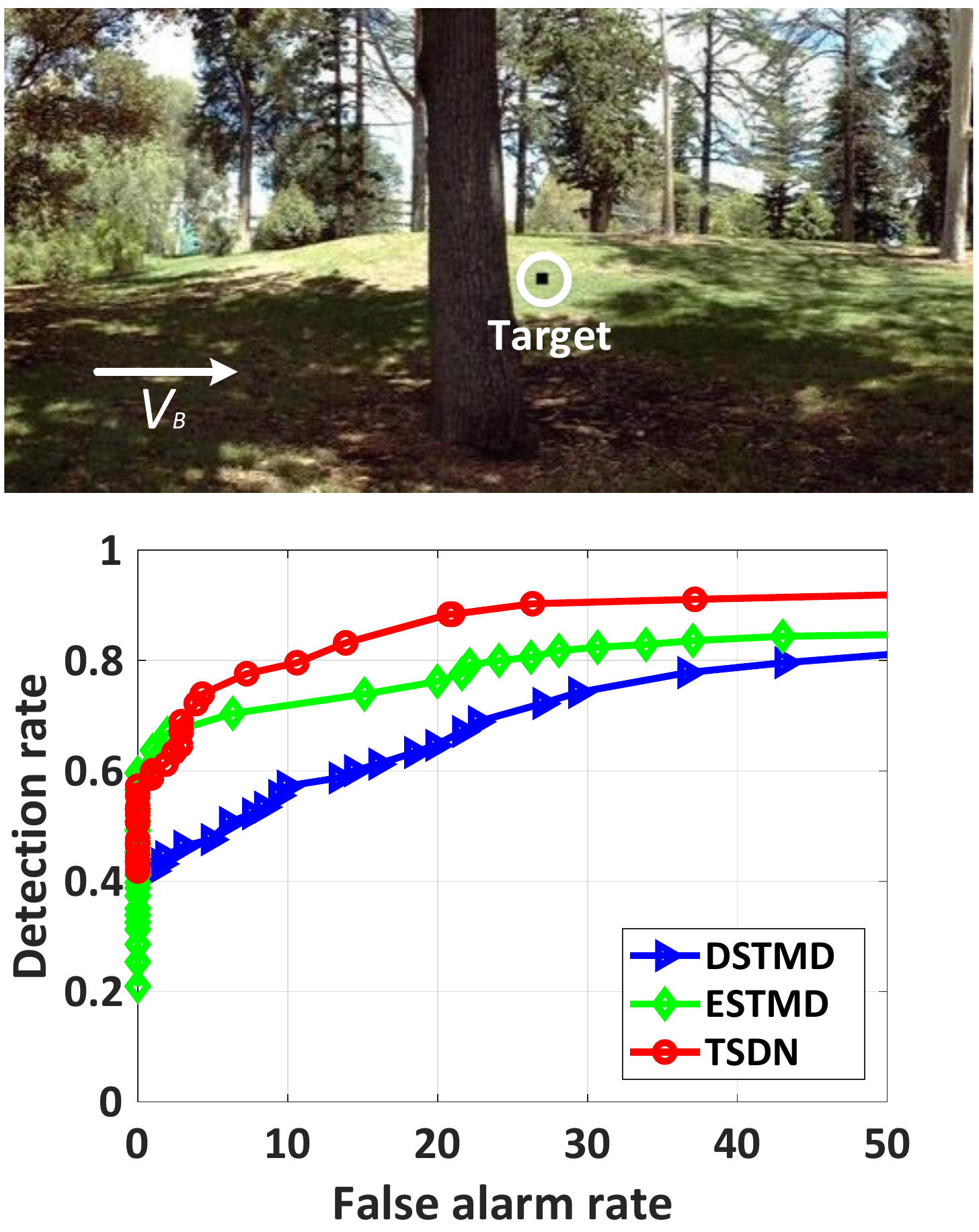}
		\label{ROC-CB-1-TDSN-DSTMD-ESTMD}}
	\hfil
	\subfloat[]{\includegraphics[width=0.3\textwidth]{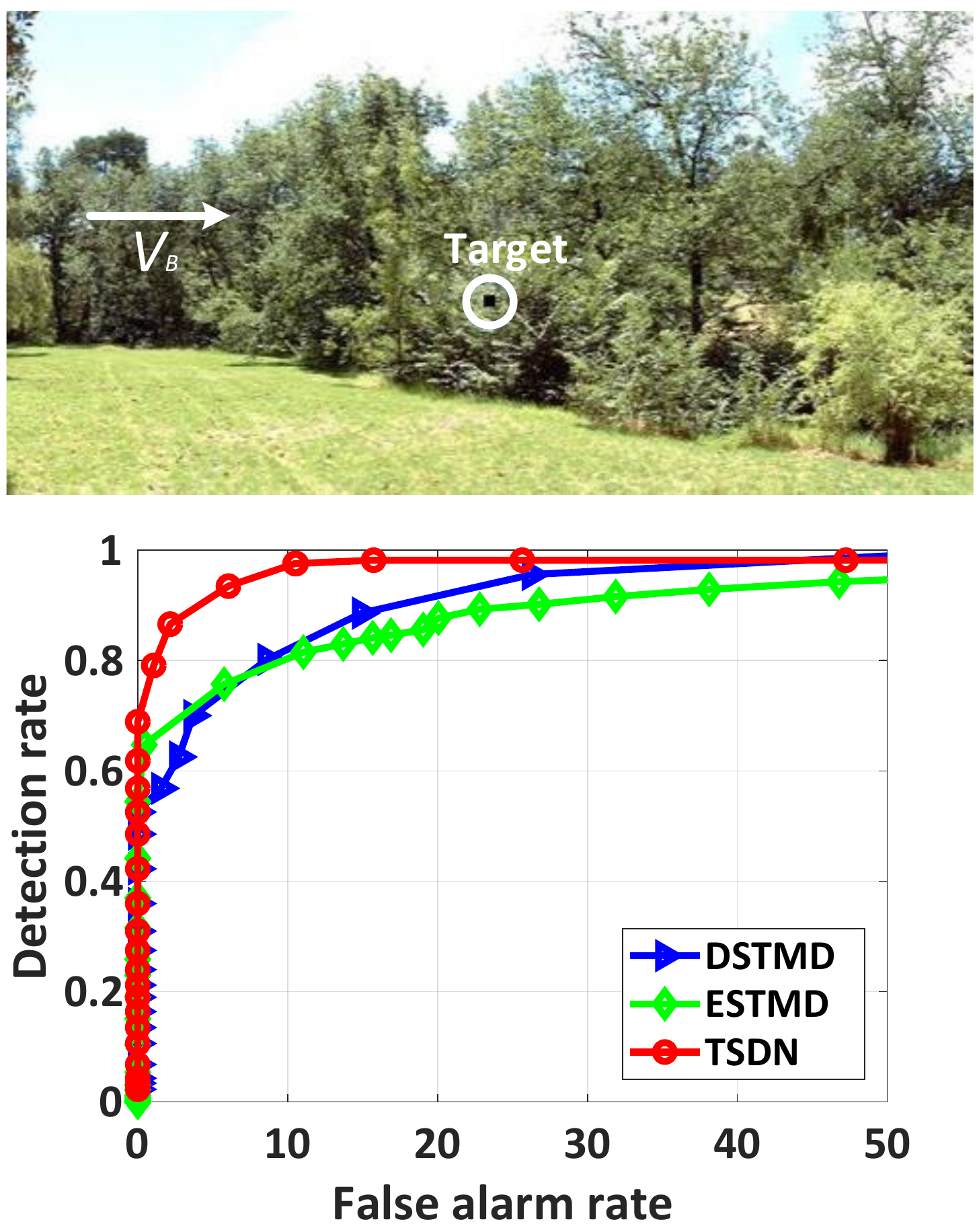}
		\label{ROC-CB-2-TDSN-DSTMD-ESTMD}}
	\hfil
	\subfloat[]{\includegraphics[width=0.3\textwidth]{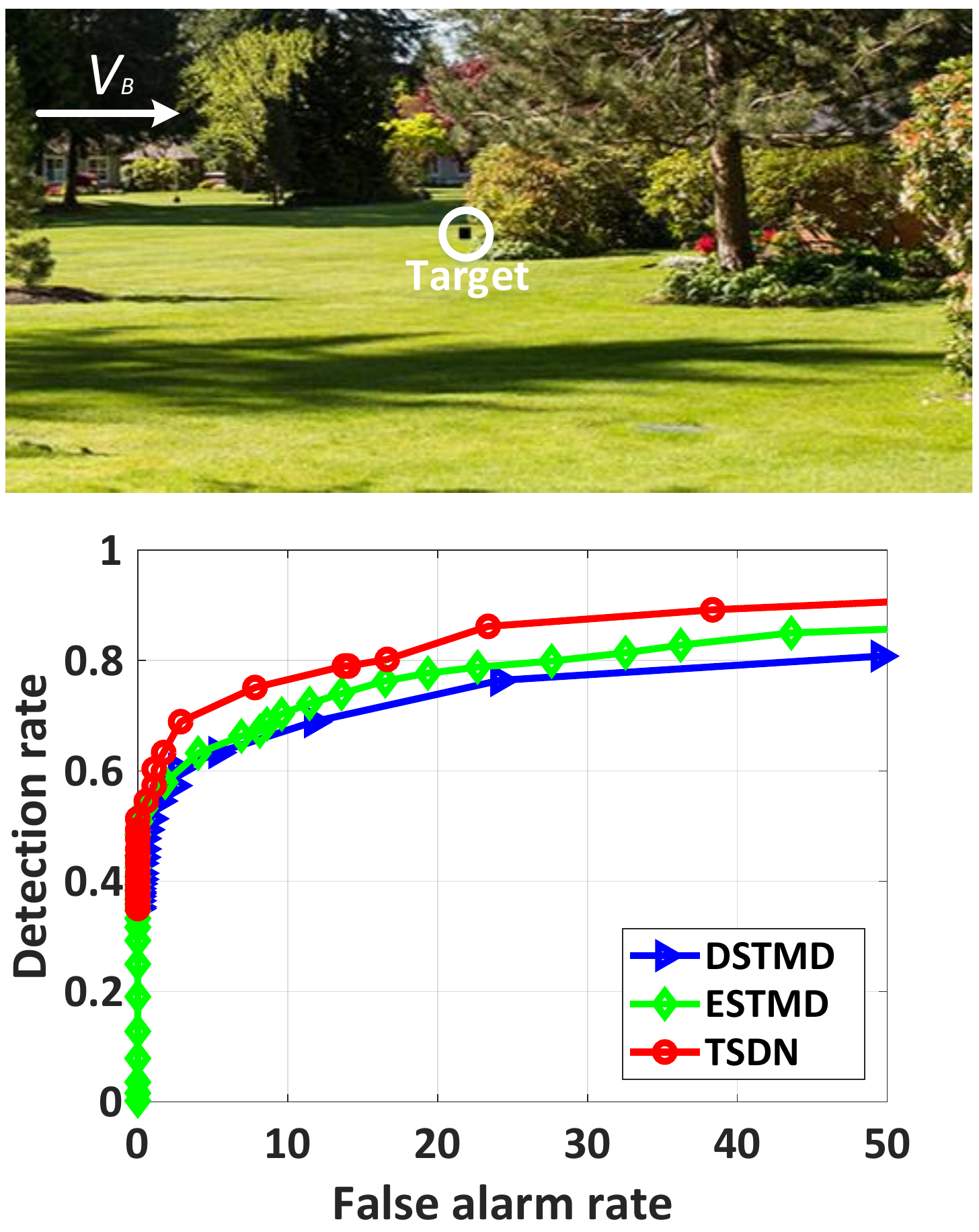}
		\label{ROC-CB-3-TDSN-DSTMD-ESTMD}}
	\caption{(a)-(b) Background images and receiver operating characteristic (ROC) curves of the three models under three different backgrounds.}
	\label{Detection-Performance-Differnet-Backgrounds}
\end{figure*}

In the proposed visual system model, the TSDNs integrate small target motion from STMDs with background motion from LPTCs to filter out background false positives. To validate its effectiveness, we compare the performances of the TSDNs and STMDs. The testing setups are detailed as follows: the input image sequence is presented in Fig. \ref{Curvilinear-Motion-Original-Image-and-Target-Trace}(a), which displays a small target moving against the cluttered background; the background is moving from left to right and its velocity is $250$ pixel/s; the luminance, size and velocity of small target are equal to $0$, $5 \times 5$ pixels and $250$ pixel/s, respectively; the position and motion direction of the small target at time $t = 1000$ ms is illustrated in Fig. \ref{Curvilinear-Motion-Original-Image-and-Target-Trace}(b)

Fig. \ref{Detection-Result-TSDN-DSTMD}(a)-(b) displays the positions and motion directions of the small objects detected by the STMDs and TSDNs where the detection threshold $\beta$ is set as $150$. As shown in Fig. \ref{Detection-Result-TSDN-DSTMD}(a), the detection result of the STMDs contains a number of false positives whose motion directions are consistent with that of the background, i.e., the blue points. After being suppressed by the output of the LPTCs, these false positives moving with the background are all filtered out [see Fig. \ref{Detection-Result-TSDN-DSTMD}(b)]. 

We further conduct a performance comparison between the developed TSDN and two existing models including ESTMD \cite{wiederman2008model} and DSTMD \cite{wang2018directionally}. Three image sequences with different backgrounds are used for experiments, as displayed in Fig. \ref{Detection-Performance-Differnet-Backgrounds}(a)-(c). In these videos,  the backgrounds are all moving from left to right and its velocity is $250$ pixel/s. A small target whose luminance, size are set as $0$ and $5 \times 5$ pixels, is moving against cluttered backgrounds. The coordinate of the small target at time $t$ is $(500 - 250 \frac{t+300}{1000}, 125+15 \sin(4\pi \frac{t+300}{1000})), t \in [0, 1000]$ ms.

Fig. \ref{Detection-Performance-Differnet-Backgrounds}(a)-(c) show the receiver operating characteristics (ROC) curves of the three models for the three image sequences. It can be seen that the TSDN has better performance than the DSTMD and ESTMD. More precisely, the TSDN has higher detection rates ($D_R$) compared to the DSTMD and ESTMD while the false alarm rates $F_A$ are low.

\section{Conclusion}
\label{Conclusion}
In this paper, we have proposed a visual system model (TSDN) for small target motion detection in cluttered backgrounds. The visual system contains two motion-sensitive neurons and is capable of filtering out the background false positives. The first neuron callled the STMD, is intended to detect small moving targets and their motion directions. The second neuron called the LPTC, is designed to perceive background motion and estimate background motion direction. The TSDN is introduced to integrate the small target motion and background motion to suppress false positives. Comprehensive evaluation on the dataset, and comparisons with the existing STMD models  demonstrate the effectiveness of the proposed visual system with lower false positives.

%

\bibliographystyle{IEEEtran}

\bibliography{IEEEabrv,Reference}

\end{document}